\documentclass{l4dc2024}
\usepackage{amsmath}  
\usepackage{amssymb}
\usepackage{mathtools}
\usepackage{caption} 
\usepackage{float}
\usepackage{comment}
\usepackage{subcaption}
\usepackage{algorithm,algcompatible,algpseudocode}
\usepackage{indentfirst} 
\newtheorem*{assumption*}{\assumptionnumber}
\providecommand{\assumptionnumber}{}
\makeatletter

\usepackage{amsfonts} 
\usepackage{graphicx} 

\DeclareMathOperator*{\argmin}{argmin}

\DeclareMathOperator*{\diag}{Diag}

\newcommand{\norm}[1]{\left\lVert#1\right\rVert}
\newcommand{\abs}[1]{\left|#1\right|}
\newtheorem*{remark_xiao}{Remark}

\newcommand{\R}{\mathbb{R}}

\newcommand{\problem}[1]{\vspace{1em}\noindent\textit{\textbf{#1}}:}

\usepackage{stmaryrd}
\newcommand{\IversonBracket}[1]{\llbracket #1 \rrbracket}

\DeclareMathAlphabet{\mymathbb}{U}{BOONDOX-ds}{m}{n}
\newcommand{\one}{\mymathbb{1}}
\newcommand{\zero}{\mymathbb{0}}

\title[System-level Safety Guard: Safe Tracking through Uncertain NN Dynamics Models]{System-level Safety Guard: Safe Tracking Control through Uncertain Neural Network Dynamics Models}
\usepackage{times}

\coltauthor{
\Name{Xiao Li} \Email{hsiaoli@umich.edu}\\
\Name{Yutong Li} \Email{yutli@umich.edu}\\
\Name{Anouck Girard} \Email{anouck@umich.edu}\\
\Name{Ilya Kolmanovsky} \Email{ilya@umich.edu}\\
\addr University of Michigan, Ann Arbor, MI, USA
}

\begin{document}

\def\arxiv{1} 
\maketitle
\begin{abstract}%
The Neural Network (NN), as a black-box function approximator, has been considered in many control and robotics applications. However, difficulties in verifying the overall system safety in the presence of uncertainties hinder the deployment of NN modules in safety-critical systems. In this paper, we leverage the NNs as predictive models for trajectory tracking of unknown dynamical systems. We consider controller design in the presence of both intrinsic uncertainty and uncertainties from other system modules. In this setting, we formulate the constrained trajectory tracking problem and show that it can be solved using Mixed-integer Linear Programming (MILP). The proposed MILP-based approach is empirically demonstrated in robot navigation and obstacle avoidance through simulations. The demonstration videos are available at \url{https://xiaolisean.github.io/publication/2023-11-01-L4DC2024}. \let\thefootnote\relax\footnotetext{This research is supported in part by AFOSR grant number FA9550-20-1-0385 and by National Science Foundation Award CMMI-1904394.}

\end{abstract}

\begin{keywords}%
neural networks, system-level safety, uncertainties, trajectory tracking%
\end{keywords}

\section{Introduction}\label{sec:intro}
Robotic and autonomous driving systems are typically structured as a pipeline of individual modules, which are designed separately to satisfy corresponding performance requirements and are verified at a system level. With recent advances in machine learning, NNs have been utilized in individual modules, e.g., localization (\cite{li2021seannet}), mapping (\cite{roddick2020predicting}), and path planning (\cite{barnes2017find}). However, the NNs approximate the desired functionalities as nonlinear mappings from data, thereby introducing (intrinsic) approximation errors. On a system level, the performance of an NN module can also be affected by extrinsic uncertainties from other modules. In safety-critical scenarios, considering both intrinsic and extrinsic uncertainties is crucial to the module design, yet vital to securing safety at the system level.

\begin{figure}[t!]
\begin{center}
\includegraphics[width=0.98\linewidth]{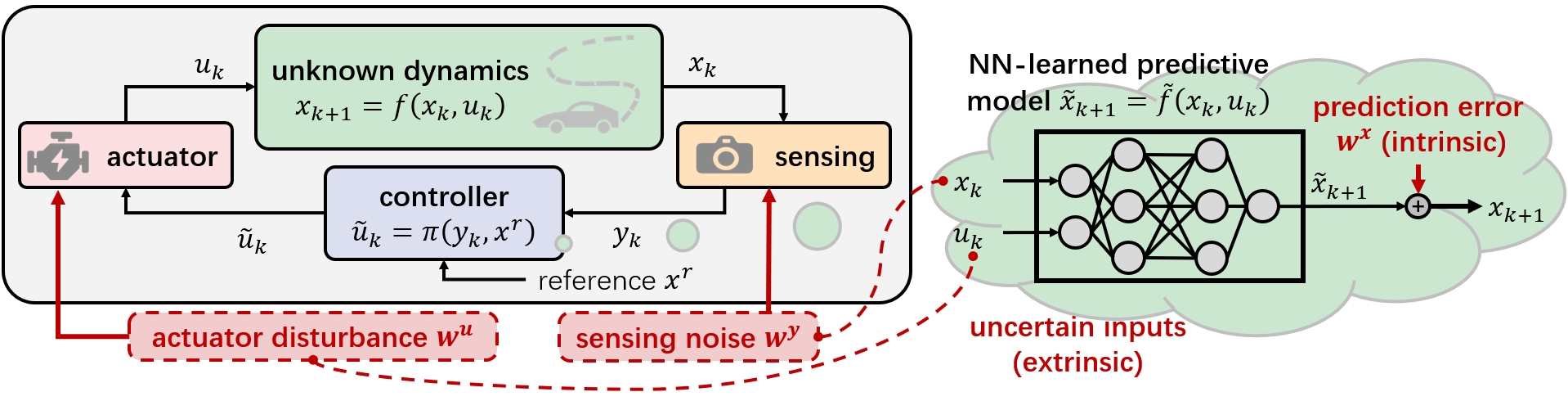}
\end{center}\vspace{-2em}
\caption{Trajectory tracking through an NN-learned predictive model: The controller leverages an NN-learned model (with prediction errors) for predicting the unknown dynamics and tracks the reference trajectories. The NN predictions depend on the states $x_k$ before the sensing module and the control $u_k$ out of the actuator module, which are not directly accessible by the controller in this pipeline, and are uncertain quantities due to sensing noises and actuator disturbances.}
\label{fig:nndm}\vspace{-1em}
\end{figure}

Several methods have been proposed to employ NNs in control development with safety guarantees. In particular, constrained nonlinear optimal control problems have been pursued to control complex dynamical systems leveraging NN-learned dynamic models, e.g., control of quadrotor motion (\cite{bansal2016learning}) and reduction of diesel engine emission (\cite{zhang2023model}). To mitigate the computation effort, methods based on MILP have been employed to embed NNs with ReLU activation functions in Model Predictive Control formulations (\cite{wei2021safe}). However, the notion of uncertainties has not been considered in the literature above. In this work, we consider trajectory tracking controller design leveraging NNs as predictive models (see Figure~\ref{fig:nndm}). Apart from NN prediction errors, we consider the presence of uncertainties from other modules, which affect the NN predictions and, subsequently, impact the controller design and the system's safety.

In addition, several methods have been investigated for the safety verification of NNs, where safety constraints are imposed on NN outputs for a fixed set of inputs. The MILP-based methods have been utilized for evaluating NN classifiers with ReLU activation functions (\cite{tjeng2017evaluating}). Approaches to reachability analysis based on Bernstein polynomials (\cite{huang2019reachnn}) and Semi-Definite Programming (SDP) (\cite{Reach-SDP}) have been proposed to estimate an overbound of the reachable set given an input domain and a NN representation. Differently from the existing literature, we employ NNs as predictive models for dynamical systems, and the NN inputs are affected by uncertainties that belong to a decision-variable-dependent set. This renders the activation status of NN neurons uncertain and dependent on the decision variables of the optimal control problem. We focus on exploiting the structure of NNs in accounting for uncertainty propagation.

The contributions of this paper are as follows: 1) We propose an approach to robust tracking controller design subject to system-level safety constraints, which leverages an NN-learned dynamic model of the system. The safety and dynamics constraints are informed by decision-variable-dependent uncertainty set propagation through NN and are handled using MILP. 2) We consider both intrinsic uncertainties from NN prediction errors and extrinsic uncertainties present in other modules and establish theoretical properties for the proposed method. 3) We illustrate the applications of the proposed approaches in simulations of collision-avoiding navigation for both an omnidirectional mobile robot and a conventional vehicle.


This paper is organized as follows: In Section~\ref{sec:problem}, we introduce the assumptions on the actual dynamics of the system as well as the NN learned dynamics, and we formulate a robust tracking problem. In Section~\ref{sec:method}, we present our method to solve the robust tracking problem, using MILP and its theoretical properties. In Section~\ref{sec:resultRobot}, we use the proposed method, in combination with a Reachability-Guided RRT algorithm, to navigate an omnidirectional robot through a maze filled with obstacles. In Section~\ref{sec:resultCar}, we leverage a set-theoretical localization algorithm that provides vehicle state measurements with uncertainty bounds, and we use our method to navigate a vehicle while avoiding collisions. Finally, conclusions are given in Section~\ref{sec:conclusion}. Due to length constraints, the proofs of the theoretical properties are relegated to 
\if\arxiv1
Appendix~\ref{sec:safety_proof}.
\else
our full technical report (\cite{techReport}).
\fi
\section{Problem Formulation}\label{sec:problem} 
We consider a discrete-time dynamical system represented by $x_{k+1} = f(x_k, u_k)$, where the state $x_k\in\mathcal{X}$ evolves within a specified feasible set $\mathcal{X}\subset\mathbb{R}^{n_x}$; $u_k\in\mathcal{U}$ is the control and $\mathcal{U}\subset\mathbb{R}^{n_u}$ is the control admissible set; and $f:\mathcal{X}\times\mathcal{U}\rightarrow\mathcal{X}$ is an unknown nonlinear mapping. In the sequel, we use lowercase letters, e.g., $x,y,u,w$, to represent vectors, capital letters, e.g., $W$, to define matrices, and scripted letters, e.g., $\mathcal{X},\mathcal{U}$, to denote sets. With a slight abuse of notation, we use $a\leq b$ to denote the element-wise order between two vectors $a,b\in\R^n$. In addition, given a lower bound $a\in\R^n$, an upper bound $b\in\R^n$ and $a\leq b$, we use $[a,b]$ to denote a hypercube in $\R^n$. We use $I_{n\times n}$ to represent the identity matrix of size $n$. We use $\one_{n\times m}$, and $\zero_{n\times m}$ to denote matrices with all zeros and ones, respectively, of size $n\times m$. We neglect the subscript $n\times m$ in calculations assuming that the dimensions are appropriate. We first discuss the NN-learned dynamic model for controller design that is subject to both intrinsic and extrinsic uncertainties in Section~\ref{subsec:model}. Then, we introduce the robust tracking problem ensuring safety under uncertainties in Section~\ref{subsec:problem}.

\subsection{Model Preliminaries}\label{subsec:model}
As shown in Figure~\ref{fig:nndm}, we approximate the dynamics $f$ with a pre-trained $\ell-$layer fully connected neural network (NN) $\tilde{f}:\mathcal{X}\times\mathcal{U}\rightarrow\mathcal{X}$ that admits the following form
\vspace{-0.5em}\begin{equation}\label{eq:nn_structure}
\begin{array}{c}
    z_i = \sigma^{(i)}(\hat{z}_i),\quad \hat{z}_i = W^{(i)}z_{i-1} +b^{(i)},\quad  i=1,\dots,\ell-1,\\
    z_0 = [x_k^T, u_k^T]^T,\quad
    \tilde{x}_{k+1} = W^{(\ell)}z_{\ell-1} +b^{(\ell)},
\end{array}
\vspace{-0.5em}\end{equation}
where $W^{(i)}\in\mathbb{R}^{n_i\times n_{i-1}}$, $b^{(i)}\in\mathbb{R}^{n_i}$ and $\sigma^{(i)}:\mathbb{R}^{n_i}\rightarrow\mathbb{R}^{n_i}$ are the weight matrix, the bias vector and an element-wise nonlinear activation function in the $i${th} layer, respectively. The NN uses the inputs $x_k,u_k$ to compute a prediction $\tilde{x}_{k+1}=\tilde{f}(x_k,u_k)$ of the actual state $x_{k+1}$. We consider the overall system to be subject to uncertainties (see Figure~\ref{fig:nndm}) of three types: Firstly, the NN prediction $\tilde{x}_{k+1}$ of the actual state $x_{k+1}$ is subject to an unknown prediction error $w^x_k$, i.e., 
\vspace{-0.5em}\begin{equation}\label{eq:x_bound}
    \tilde{x}_{k+1} = x_{k+1} + w^x_k,\;w^x_k\in \mathcal{W}_x,
\vspace{-0.5em}\end{equation}
where $\mathcal{W}_x\subset \mathbb{R}^{n_x}$ is a bounded set; moreover, the actual states $x_k$ and the actual control input $u_k$ are both unknown to the controller due to the presence of noises in other modules. Secondly, we assume that a state measurement $y_k$ of the actual state $x_k$ is available from the sensing module and admits the form,
\vspace{-0.5em}\begin{equation}\label{eq:y_bound}
    y_k = x_k + w^y_k,\;w^y_k\in \mathcal{W}_y,
\end{equation}
where $w^y_k$ is an unknown measurement noise, and $\mathcal{W}_y\subset \mathbb{R}^{n_x}$ is a bounded set. Thirdly, we consider the use of a feedback controller $\pi:\mathcal{X}\times \mathcal{X}\to\mathcal{U}$ to track the reference state $x^r$ (to be designed). Then the actual control signal $u_k$ is subject to an unknown additive actuator disturbance $w^u_k$, i.e., 
\vspace{-0.5em}\begin{equation}\label{eq:u_bound}
    u_k = \tilde{u}_k + w^u_k,\;\tilde{u}_k=\pi(y_k, x^r), \;w^u_k\in \mathcal{W}_u,
\vspace{-0.5em}\end{equation}
where $\mathcal{W}_u\subset \mathbb{R}^{n_u}$ is a bounded set.

\subsection{Robust Constrained Tracking Problem}\label{subsec:problem}
Assume that $\mathcal{X}_u\subset\mathcal{X}$ is an unsafe set, e.g., representing obstacles, that the system should avoid. Given a reference $x^r\in\mathcal{X}_s$ in the safe subset, $\mathcal{X}_s=\mathcal{X}\backslash\mathcal{X}_u$, and a state measurement $y_k$ that obeys the assumption in~\eqref{eq:y_bound}, our task is to design a controller $\pi(y_k, x^r)$ to track the reference $x^r$ while keeping the system in the safe subset $\mathcal{X}_s$, i.e., ensuring $x_{k+1}\in \mathcal{X}_s$ given $x_k\in \mathcal{X}_s$. These objectives can be accounted for in a constrained optimization problem,
\begin{subequations}\label{eq:safeOptProblem}
\begin{equation}\label{eq:OP_objective}
    \pi(y_k, x^r) = \argmin\limits_{\tilde{u}_k}{\Big(\max\limits_{x\in\mathcal{F}\big(\mathcal{X}_k,\;\mathcal{U}_k(\tilde{u}_k)\big)}\norm{x-x^r}_p\Big)}
\end{equation}
\begin{equation}\label{eq:OP_input}
     \text{subject to: \quad}\tilde{u}_k\in\mathcal{U},\; \mathcal{X}_k = \big(y_k\oplus-\mathcal{W}_y\big)\cap\mathcal{X}_s,\;\mathcal{U}_k=\mathcal{U}_k(\tilde{u}_k) = (\tilde{u}_k\oplus\mathcal{W}_u)\cap\mathcal{U},\quad\quad\quad\quad
\end{equation}
\begin{equation}\label{eq:OP_structure}
\begin{array}{c}
    \tilde{\mathcal{F}}(\mathcal{X}_k,\mathcal{U}_k)=\left\{\tilde{x}_{k+1}\in\mathcal{X}:\; \tilde{x}_{k+1}=\tilde{f}(x_k,u_k),\forall x_k\in\mathcal{X}_k, \forall u_k\in\mathcal{U}_k\right\},
\end{array}
\end{equation}
\begin{equation}\label{eq:OP_output}
    \mathcal{F}(\mathcal{X}_k,\mathcal{U}_k) =\tilde{\mathcal{F}}(\mathcal{X}_k,\mathcal{U}_k)\oplus\mathcal{W}_x,\;
    \mathcal{F}(\mathcal{X}_k,\mathcal{U}_k)\subset\mathcal{X}_s,
\end{equation}
\end{subequations}
where $\tilde{u}_k$ is the decision variable, $\norm{\cdot}_p$ represents the $\ell_p$ norm, and $\oplus$ denotes the Minkowski sum. The term $y_k\oplus-\mathcal{W}_y$ stands for $\{y_k\}\oplus\{w:-w\in\mathcal{W}_y\}$ where we omit the curly brackets for simplicity. Given the measurement $y_k$ that satisfies \eqref{eq:y_bound}, the set $\mathcal{X}_k$ in constraint~\eqref{eq:OP_input} represents an uncertainty set that contains the actual state $x_k$. Similarly, the set $\mathcal{U}_k$ in constraint~\eqref{eq:OP_input} depends on the decision variable $\tilde{u}_k$ and contains the actual control $u_k$ based on \eqref{eq:u_bound}. The set-valued function $\tilde{\mathcal{F}}:\mathcal{X}\times\mathcal{U}\rightarrow\mathcal{X}$ in~\eqref{eq:OP_structure} calculates the one step ahead reachable set of the learned NN dynamic model. The reachable set $\mathcal{F}(\mathcal{X}_k,\mathcal{U}_k)$ in \eqref{eq:OP_output} is  derived from $\tilde{\mathcal{F}}(\mathcal{X}_k,\mathcal{U}_k)$ and contains the actual state $x_{k+1}$ based on \eqref{eq:x_bound}. Meanwhile, the constraints in~\eqref{eq:OP_output} also guarantee the safety of the next state $x_{k+1}$ for all the possible uncertainties modeled by~\eqref{eq:x_bound},~\eqref{eq:y_bound},~\eqref{eq:u_bound}. The objective defined by \eqref{eq:OP_objective} minimizes the maximum distance between the states in the reachable set $\mathcal{F}(\mathcal{X}_k,\mathcal{U}_k)$ and the reference, thereby steering the system closer to $x^r$. Unlike reachability analysis~\cite{Reach-SDP}, the input set $\mathcal{U}_k=\mathcal{U}_k(\tilde{u}_k)$ is conditioned on the decision variable $\tilde{u}_k$. 

\begin{remark_xiao}
We introduce our methodology in the simplest possible setting, i.e., horizon-one Model Predictive Control with no control cost. This is the same setting as considered in \cite{wei2021safe}. The methodology can be extended to accommodate a multi-step prediction horizon, by augmenting \eqref{eq:OP_structure} and \eqref{eq:OP_output} with the following constraints:
\vspace{-0.5em}\begin{equation*}
    \mathcal{F}^{(i+1)} = \tilde{\mathcal{F}}^{(i+1)}\oplus\mathcal{W}_x\subset\mathcal{X}_s,\; 
    \tilde{\mathcal{F}}^{(i+1)} = \tilde{\mathcal{F}}(\mathcal{F}^{(i)},\mathcal{U}_k),\;
    i = 1,2,\dots, N-1, 
\vspace{-0.5em}\end{equation*}
where $N$ is the prediction horizon length, and $\mathcal{F}^{(1)}=\mathcal{F}(\mathcal{X}_k,\mathcal{U}_k)$. We consider $p=1$ in the sequel as it leads to linear constraints and MILP.  
\end{remark_xiao}

\section{Mixed-integer Linear Programming}\label{sec:method}
To simplify the exposition of the approach, we assume that the uncertainty sets in \eqref{eq:x_bound}, \eqref{eq:y_bound}, \eqref{eq:u_bound} are hyper-cubes defined according to\vspace{-0.5em}
\begin{subequations}\label{eq:bound_cube}
\begin{equation}\label{eq:x_bound_cube}
    \mathcal{W}_x = \left\{w^x\in\mathbb{R}^{n_x}: \abs{w^x}\leq\epsilon^x,\;\epsilon^x\in\mathbb{R}^{n_x},\;\epsilon^x\geq0\right\},
\end{equation}
\begin{equation}\label{eq:y_bound_cube}
    \mathcal{W}_y = \left\{w^y\in\mathbb{R}^{n_x}: \abs{w^y}\leq\epsilon^y,\;\epsilon^y\in\mathbb{R}^{n_x},\;\epsilon^y\geq0\right\},
\end{equation}
\begin{equation}\label{eq:u_bound_cube}
    \mathcal{W}_u = \left\{w^u\in\mathbb{R}^{n_u}: \abs{w^u}\leq\epsilon^u,\;\epsilon^u\in\mathbb{R}^{n_u},\;\epsilon^u\geq0\right\}.
\end{equation}
\end{subequations}
Note that, in principle, one can always find a hypercube overbounding a bounded set. Moreover, we assume that the nonlinear functions in the NN model are ReLU activation functions
\vspace{-0.5em}\begin{equation}\label{eq:relu}
    \sigma^{(i)}(x) = {\tt ReLU} (x)= \max\{0,x\},\;i=1,\cdots,\ell,
\vspace{-0.5em}\end{equation}
that are commonly adopted in contemporary NN architectures and have demonstrated good empirical performance (\cite{lecun2015deep}). We, furthermore, focus on the case when the state feasible set, control admissible set, and unsafe set are represented as
\vspace{-0.5em}
\begin{subequations}\label{eq:feasible_set}
\begin{equation}\label{eq:x_feasible_set}
    \mathcal{X} = [\underline{x}, \overline{x}],\;\underline{x},\overline{x}\in\mathbb{R}^{n_x},\;\underline{x}\leq\overline{x},
\end{equation}
\begin{equation}\label{eq:u_feasible_set}
    \mathcal{U} = [\underline{u}, \overline{u}],\;\underline{u},\overline{u}\in\mathbb{R}^{n_u},\;\underline{u}\leq\overline{u},
\end{equation}
\begin{equation}\label{eq:x_unsafe_set}
    \mathcal{X}_u=\cup_i \mathcal{X}_u^{(i)},\; 
    \mathcal{X}_u^{(i)} = [\underline{x}_u^{(i)}, \overline{x}_u^{(i)}],\;\underline{x}_u^{(i)},\overline{x}_u^{(i)}\in\mathbb{R}^{n_x},\;\underline{x}\leq\underline{x}_u^{(i)}\leq\overline{x}_u^{(i)}\leq\overline{x}.
\end{equation}
\end{subequations}
The definition of the unsafe set~\eqref{eq:x_unsafe_set} enables us to tightly over-bound obstacles of irregular shapes using unions of hypercubes in the optimization problem. Then, equations \eqref{eq:OP_input}, \eqref{eq:OP_structure}, \eqref{eq:OP_output} encode the constraints associated with the input feasibility, NN structural non-linearity, and system safety, respectively, and are realized using integer decision variables as described in the respective Sections \ref{subsec:constraint_input}, \ref{subsec:constraint_nn}, and \ref{subsec:constraint_output}. In this setting and when $p=1$ in \eqref{eq:OP_objective}, i.e., the cost function is based on the $\ell_1$ norm, we show that \eqref{eq:safeOptProblem} reduces to a MILP in Section~\ref{subsec:constraint_obj}. The proofs of the theoretical properties in this section are provided in 
\if\arxiv1
Appendix~\ref{sec:safety_proof}.
\else
our full technical report (\cite{techReport}).
\fi


\subsection{Constraints Embedding Input Feasibility}\label{subsec:constraint_input}
Given a measurement $y_k$ and the decision variable $\tilde{u}_k$, the constraints~\eqref{eq:OP_input} define the sets $\mathcal{X}_k$, $\mathcal{U}_k$ that are guaranteed to contain the actual state $x_k$ and the control $u_k$ while the decision variable shall be admissible, i.e., $x_k\in\mathcal{X}_k$, $u_k\in\mathcal{U}_k$, $\tilde{u}_k\in\mathcal{U}_k$. We embed these feasibility conditions of the NN inputs using linear inequality and equality constraints in the following proposition. 

\begin{proposition}\label{prop:constraint_input}
Given state measurement $y_k$ and a decision variable $\tilde{u}_k$, assume that the unknown actual quantities $x_k$ and $u_k$ obey \eqref{eq:y_bound}, \eqref{eq:u_bound} with assumptions in \eqref{eq:y_bound_cube}, \eqref{eq:u_bound_cube}, \eqref{eq:x_feasible_set}, \eqref{eq:u_feasible_set}. Let the decision variables $\tilde{u}_k \in \R^{n_u}$, $a_0,b_0\in\R^{n_x+n_u}$, and $\delta^a, \delta^b\in\R^{n_u}$ satisfy the following constraints
\begin{equation}\label{eq:mip_input_x}
    a_{0,1 : n_x} = \max\{\underline{x}, y_k-\epsilon^y\},\quad
    b_{0,1 : n_x} = \min\{\overline{x}, y_k+\epsilon^y\},\quad
    a_0 \leq b_0,
\end{equation}
\begin{equation}\label{eq:mip_input_u}
\begin{array}{c}
    \underline{u}\leq \tilde{u}_k \leq \overline{u},\quad 
    \delta^a_j,\delta^b_j\in\{0,1\},\; j =1,\dots, n_u,
    \vspace{0.3em}\\
    \left\{\begin{array}{l}
        a_{0,(n_x+1) : (n_x+n_u)} \geq \underline{u}\\
        a_{0,(n_x+1) : (n_x+n_u)} \geq \tilde{u}_k-\epsilon^u\\
        a_{0,(n_x+1) : (n_x+n_u)} \leq \underline{u} + M(\one-\delta^a)\\
        a_{0,(n_x+1) : (n_x+n_u)} \leq \tilde{u}_k-\epsilon^u + M\delta^a
    \end{array}\right.,
    \left\{\begin{array}{l}
        b_{0,(n_x+1) : (n_x+n_u)} \leq \overline{u}\\
        b_{0,(n_x+1) : (n_x+n_u)} \leq \tilde{u}_k+\epsilon^u\\
        b_{0,(n_x+1) : (n_x+n_u)} \geq \overline{u} - M(\one-\delta^b)\\
        b_{0,(n_x+1) : (n_x+n_u)} \geq \tilde{u}_k+\epsilon^u - M\delta^b
    \end{array}\right.,
\end{array}
\end{equation}
where $\delta^a_j$ denotes the $j${th} element in column vector $\delta^a$, $a_{0,m:n}$ represents the vector containing elements between row $n$ and row $m$ in $a_0$, the constant matrix $M=\diag\left(\max\{\epsilon^u, \overline{u}-\underline{u}-\epsilon^u\}\right)$,  and $\diag(x)\in\R^{n\times n}$ yields a square matrix with elements of $x$ on the diagonal and zero anywhere else. Then, it is guaranteed that $z_0=[x_k^T\; u_k^T]^T\in\left[a_0,\;b_0\right]$.
\end{proposition}

The constraints in Proposition~\ref{prop:constraint_input} enforce input feasibility, i.e.,  $\tilde{u}_k\in\mathcal{U}_k$, and provide bounds $a_0,b_0$ on the NN input subject to extrinsic uncertainties, i.e., $a_0\leq [x^T,\;u^T]^T\leq b_0$ for all $x\in\mathcal{X}_k,\; u\in\mathcal{U}_k$. Specifically, based on assumption \eqref{eq:y_bound_cube} and \eqref{eq:x_feasible_set}, the constraints \eqref{eq:mip_input_x} imply $\mathcal{X}_k \subseteq [a_{0,1 : n_x}, b_{0,1 : n_x}]$. Based on assumption \eqref{eq:u_bound_cube} and \eqref{eq:u_feasible_set}, the constraints \eqref{eq:mip_input_u} are equivalent to the following inequalities
\vspace{-0.5em}\begin{equation*}
    \max\{\underline{u}, \tilde{u}_k-\epsilon^u\}= 
    a_{0,(n_x+1)\cdots (n_x+n_u)}
    \leq 
    b_{0,(n_x+1)\cdots (n_x+n_u)}
    = \min\{\overline{u}, \tilde{u}_k+\epsilon^u\},
\vspace{-0.5em}\end{equation*}
thereby $\mathcal{U}_k\subseteq [a_{0,(n_x+1) : (n_x+n_u)}, b_{0,(n_x+1) : (n_x+n_u)}]$. We use integer variables $\delta^a,\delta^b$ to move the decision variable $\tilde{u}_k$ out of the nonlinear min/max function: $\delta^a_j=1$ implies the $j${th} element of $\max\{\underline{u}, \tilde{u}_k-\epsilon^u\}$ attains the value of the $j${th} element of $\underline{u}$, otherwise $\delta^a_j=0$; $\delta^b_j=1$ indicates the $j${th} element of $\min\{\overline{u}, \tilde{u}_k+\epsilon^u\}$ attains the value of the $j${th} element of $\overline{u}$, otherwise $\delta^b_j=0$. 
\if\arxiv1
The proof is presented in \ref{subsec:safety_proof_input}.
\fi
\subsection{Constraints Encoding NN Structural Non-Linearity}\label{subsec:constraint_nn}
Using the bounded sets  $\mathcal{X}$ in \eqref{eq:x_feasible_set} and $\mathcal{U}$ in \eqref{eq:u_feasible_set}, we can numerically derive lower and upper bounds $\underline{\hat{z}}_i,\overline{\hat{z}}_i\in\mathbb{R}^{n_i},i=1,\dots,\ell$ on the neuron values $\hat{z}_i$ and the output $\tilde{x}_{k+1}$ using interval arithmetic, i.e., $\hat{z}_i\in[\underline{\hat{z}}_i,\overline{\hat{z}}_i]$ and $\tilde{x}_{k+1}\in[\underline{\hat{z}}_{\ell},\overline{\hat{z}}_{\ell}]$. In the sequel, these derived bounds are used to tighten the constraints and limit the search region for the optimization solver. Given the decision variables $a_0,b_0$ in Proposition~\ref{prop:constraint_input} as bounds on the NN input, we can encode the decision-variable-dependent uncertainty set propagation through the NN defined in \eqref{eq:OP_structure} using the following results:
\begin{proposition}\label{prop:constraint_nn}
Given $z_0=[x_k^T\; u_k^T]^T\in\left[a_0,\;b_0\right]$, consider a NN defined by \eqref{eq:nn_structure} and \eqref{eq:relu}, and let the decision variables $a_{i-1},b_{i-1}\in\mathbb{R}^{n_{i-1}}$, $\hat{a}_i,\hat{b}_i\in\mathbb{R}^{n_i}$, $\delta^{--}_{i}, \delta^{-+}_{i}, \delta^{++}_{i}\in\{0,1\}^{n_i}$, $i=1,\dots,\ell-1$, and $a_{k+1}, b_{k+1}\in\mathbb{R}^{n_x}$ satisfy the following constraints
\begin{equation}\label{eq:mip_struct_linear}
\begin{array}{c}
    \hat{a}_{i,j}=w^{(i)}_j S_i\left((w^{(i)}_j)^T\right)
    \left[\begin{array}{c}
        a_{i-1}\\
        b_{i-1}
    \end{array}\right]
    + b^{(i)}_j
    ,\quad
    \hat{b}_{i,j}=w^{(i)}_j S_i\left((w^{(i)}_j)^T\right)
    \left[\begin{array}{c}
        b_{i-1}\\
        a_{i-1}
    \end{array}\right]
    + b^{(i)}_j,
    \vspace{0.3em}\\
    \underline{\hat{z}}_i\leq\hat{a}_i\leq\hat{b}_i\leq\overline{\hat{z}}_i,
    \quad
    \forall j=1,\cdots,n_i,\quad \forall i=1,\dots,\ell-1,
\end{array}    
\end{equation}
\begin{equation}\label{eq:mip_struct_relu}
\begin{array}{c}
    \left\{\begin{array}{l}
        a_i \geq \hat{a}_i\\
        a_i \leq \hat{a}_i - \diag(\underline{\hat{z}}_i)(\delta^{--}_i+\delta^{-+}_i)\\
        a_i \leq \diag(\overline{\hat{z}}_i)\delta^{++}_i
    \end{array}\right.,
    \quad
    \left\{\begin{array}{l}
        b_i \geq \hat{b}_i\\
        b_i \leq \hat{b}_i - \diag(\underline{\hat{z}}_i)\delta^{--}_i\\
        b_i \leq \diag(\overline{\hat{z}}_i)(\delta^{-+}_i+\delta^{++}_i)
    \end{array}\right.,
    \vspace{0.3em}\\
    0 \leq a_i \leq b_i,
    \quad
    \delta^{--}_{i,j}, \delta^{-+}_{i,j}, \delta^{++}_{i,j}\in\{0,1\},
    \quad
    \delta^{--}_{i,j}+ \delta^{-+}_{i,j}+\delta^{++}_{i,j} = 1,
    \\
    \forall j=1,\cdots,n_i,
    \quad 
    \forall i=1,\dots,\ell-1,
\end{array}
\end{equation}
\begin{equation}\label{eq:mip_struct_out}
\begin{array}{c}
    \underline{\hat{z}}_{\ell}\leq a_{k+1}\leq b_{k+1}\leq\overline{\hat{z}}_{\ell}
    ,\quad
    a_{k+1,j}=w^{(\ell)}_j S_{\ell}\left((w^{(\ell)}_j)^T\right)
    \left[\begin{array}{c}
        a_{\ell-1}\\
        b_{\ell-1}
    \end{array}\right]
    + b^{(\ell)}_j
    ,\\
    b_{k+1,j}=w^{(\ell)}_j S_{\ell}\left((w^{(\ell)}_j)^T\right)
    \left[\begin{array}{c}
        b_{\ell-1}\\
        a_{\ell-1}
    \end{array}\right]
    + b^{(\ell)}_j,
    \quad 
    \forall j=1,\cdots,n_i,
\end{array}    
\end{equation}
where $\hat{a}_{i,j}$ is the $j${th} element of $\hat{a}_i$; $w^{(i)}_j$ is the $j${th} row of $W^{(i)}$; $b^{(i)}_j$ is the $j${th} element of $b^{(i)}$; $\delta^{--}_{i,j}$ denotes the $j${th} element of $\delta^{--}_{i}$; $a_{k+1,j},b_{k+1,j}$ represent the $j${th} element of $a_{k+1},b_{k+1}$, respectively; The functions $S_i:\mathbb{R}^{n_{i-1}}\rightarrow\mathbb{R}^{n_{i-1}\times2n_{i-1}}$ and $s: \mathbb{R}\rightarrow\mathbb{R}^{2n_{i-1}}$ are defined according to 
\vspace{-0.5em}\begin{equation*}
\begin{array}{c}
    S_i
    \left(\left[\begin{array}{c}
        \vdots\\
        w_q\\
        \vdots
    \end{array}\right]\right)
    =
    \left[\begin{array}{c}
        \vdots\\
        \left(s(w_q)\right)^T\\
        \vdots
    \end{array}\right],
    \quad
    s(w_q) :=
    \left\{\begin{array}{cc}
        \left[\begin{array}{c}
            e_q\\
            \zero_{n_{i-1}\times 1}
        \end{array}\right]
        & \text{if } w_q \geq 0 
        \\
        \left[\begin{array}{c}
            \zero_{n_{i-1}\times 1}\\
            e_q
        \end{array}\right]
        & \text{if } w_q < 0 
    \end{array}\right.,
\end{array}
\vspace{-0.5em}\end{equation*}
and $e_q\in\mathbb{R}^{n_{i-1}}$ ($q= 1,\dots,n_{i-1}$) has 1 as the $q${th} element and other elements being zero. Then, the reachable set $\tilde{\mathcal{F}}(\mathcal{X}_k,\mathcal{U}_k)$ defined in \eqref{eq:OP_structure} is a subset of the hypercube $[a_{k+1},b_{k+1}]$, i.e., $\tilde{\mathcal{F}}(\mathcal{X}_k,\mathcal{U}_k)\subseteq [a_{k+1},b_{k+1}]$.
\end{proposition}

From the $(i-1)${th} to $i${th} layer of the NN, the uncertainty set propagation is realized through variables $a_{i-1},b_{i-1}$ and $\hat{a}_i,\hat{b}_i$ that are the lower and upper bounds of $z_{i-1}$ and $\hat{z}_i$, respectively (i.e., $z_{i-1}\in[a_{i-1}, b_{i-1}]$ and $\hat{z}_i\in[\hat{a}_i, \hat{b}_i]$). The constraints \eqref{eq:mip_struct_linear} and \eqref{eq:mip_struct_out} encode the uncertainty propagation through the fully connected layers, $\hat{z}_i = W^{(i)}z_{i-1} +b^{(i)}$ and $\tilde{x}_{k+1} = W^{(\ell)}z_{\ell-1} +b^{(\ell)}$, respectively. The uncertainty set propagation through the nonlinear ReLU function is enforced in constraints \eqref{eq:mip_struct_relu}. In constraints \eqref{eq:mip_struct_linear} and \eqref{eq:mip_struct_out}, the $q${th} row of matrix $S_i((w^{(i)}_j)^T)$ switches the upper and lower bounds of $z_{i-1}$ if the $q${th} element in $w^{(i)}_j$ is negative. Meanwhile, since the values $z_{i-1}$ in neurons fall into a bounded hypercube $[a_{i-1}, b_{i-1}]$, the activation status of each ReLU activation function is uncertain. Inspired by~\cite{tjeng2017evaluating}, we introduce integer variables $\delta^{--}_{i}, \delta^{-+}_{i}, \delta^{++}_{i}$ in constraints \eqref{eq:mip_struct_relu} to encode the uncertainty in the ReLU activation status according to
\vspace{-0.5em}\begin{equation*}
    \hat{a}_{i,j}\leq \hat{b}_{i,j}\leq 0 \text{ if } \delta^{--}_{i,j} = 1;\quad
    \hat{a}_{i,j}\leq 0 \leq\hat{b}_{i,j} \text{ if } \delta^{-+}_{i,j} = 1;\quad
    0\leq\hat{a}_{i,j}\leq \hat{b}_{i,j} \text{ if } \delta^{++}_{i,j} = 1.
\vspace{-0.5em}\end{equation*}
Furthermore, as can be shown, the hypercube defined by lower bound $a_{k+1}$ and upper bound $b_{k+1}$ is an over-estimation of the actual reachable set $\tilde{\mathcal{F}}$, i.e., $\tilde{\mathcal{F}}\subseteq [a_{k+1},b_{k+1}]$. 
\if\arxiv1
The proof is presented in Appendix~\ref{subsec:safety_proof_nn}.
\fi

\subsection{Constraints Enforcing System Safety}\label{subsec:constraint_output}
Given a hypercube $[a_{k+1},b_{k+1}]$ as an over-estimation of the reachable set $\tilde{\mathcal{F}}$ in Proposition~\ref{prop:constraint_nn}, we enforce the safety constrains \eqref{eq:OP_output} of the system such that $x_{k+1}\in\mathcal{F}(\mathcal{X}_k,\mathcal{U}_k)\subset \mathcal{X}_s$. We consider the presence of a simple unsafe subset $ \mathcal{X}_u = [\underline{x}_u, \overline{x}_u]$ in the following result and discuss the extension to a union of $\mathcal{X}_u^{(i)}$ defined in \eqref{eq:x_unsafe_set} at the end of this section.

\begin{proposition}\label{prop:constraint_output}
Given $\tilde{\mathcal{F}}(\mathcal{X}_k,\mathcal{U}_k)\subseteq [a_{k+1},b_{k+1}]$ and state space defined in \eqref{eq:x_feasible_set}, assume that the NN predictions are subject to bounded additive errors defined in \eqref{eq:x_bound} and \eqref{eq:x_bound_cube}, and the decision variables $\underline{x}_{k+1},\overline{x}_{k+1}\in\mathbb{R}^{n_x}$, $\delta^u_{1}, \delta^u_{2}\in\mathbb{R}^{n_x}$ satisfy the following constraints:
\begin{equation}\label{eq:mip_actual_reachable}
    \underline{x}\leq \underline{x}_{k+1}\leq \overline{x}_{k+1}\leq \overline{x},
    \quad
    \underline{x}_{k+1} = a_{k+1} - \epsilon^x,
    \quad
    \overline{x}_{k+1} = b_{k+1} + \epsilon^x,
\end{equation}
\begin{equation}\label{eq:mip_out_unsafe}
\begin{array}{c}
    \left\{\begin{array}{l}
        \overline{x}_{k+1} \leq \overline{x} + \diag\left(\underline{x}_u-\overline{x}\right)\delta^u_1\\
        \overline{x}_{k+1} \geq \underline{x}_u - \diag\left(\underline{x}_u-\underline{x}\right)\delta^u_1\\
        \underline{x}_{k+1} \geq \underline{x} + \diag\left(\overline{x}_u-\underline{x}\right)\delta^u_2\\
        \underline{x}_{k+1} \leq \overline{x}_u - \diag\left(\overline{x}_u-\overline{x}\right)\delta^u_2\\
    \end{array}\right.,   \quad
    \begin{array}{c}
    \delta^u_{1,j}, \delta^u_{2,j}\in\{0,1\},\quad \forall j=1,\dots,n_x,
    \\
    \delta^u_{1,j} + \delta^u_{2,j}\leq 1,\quad \forall j=1,\dots,n_x,
    \\
    \sum\limits_{j=1}^{n_x}\left(\delta^u_{1,j} + \delta^u_{2,j}\right)\geq 1,
    \end{array}
\end{array}
\end{equation}
where $\delta^u_{1,j}$, $\delta^u_{2,j}$ are the $j${th} element of $\delta^u_{1}, \delta^u_{2}$, respectively. Then, $x_{k+1}\in\mathcal{F}(\mathcal{X}_k,\mathcal{U}_k)$ where $\mathcal{F}(\mathcal{X}_k,\mathcal{U}_k)$ is the reachable set defined in \eqref{eq:OP_output}, and $\mathcal{F}(\mathcal{X}_k,\mathcal{U}_k) \subseteq [\underline{x}_{k+1}, \overline{x}_{k+1}] \subset\mathcal{X}_s$ with $\mathcal{X}_u = [\underline{x}_u, \overline{x}_u]$.
\end{proposition}

In Proposition~\ref{prop:constraint_output}, the constraints \eqref{eq:mip_actual_reachable} are equivalent to $ [\underline{x}_{k+1}, \overline{x}_{k+1}]= [a_{k+1}, b_{k+1}]\oplus\mathcal{W}_x$ and $[\underline{x}_{k+1}, \overline{x}_{k+1}]\subset \mathcal{X}$. Considering the result $\tilde{\mathcal{F}}(\mathcal{X}_k,\mathcal{U}_k)\subseteq [a_{k+1},b_{k+1}]$ from Proposition~\ref{prop:constraint_nn}, it is obvious that $\mathcal{F}(\mathcal{X}_k,\mathcal{U}_k)\subseteq [\underline{x}_{k+1}, \overline{x}_{k+1}]$ based on the definition of $\mathcal{F}(\mathcal{X}_k,\mathcal{U}_k)$ in \eqref{eq:OP_output}. Subsequently, the safety constraint of $\mathcal{F}(\mathcal{X}_k,\mathcal{U}_k)\subset\mathcal{X}_s$ can be enforced with $[\underline{x}_{k+1}, \overline{x}_{k+1}]\subset\mathcal{X}_s$, which is equivalent to $[\underline{x}_{k+1}, \overline{x}_{k+1}]\subset\mathcal{X}$ and $[\underline{x}_{k+1}, \overline{x}_{k+1}]\cap\mathcal{X}_u = \varnothing$. The constraints \eqref{eq:mip_out_unsafe} enforce $[\underline{x}_{k+1}, \overline{x}_{k+1}]\cap\mathcal{X}_u = \varnothing$ using integer variables according to 
\vspace{-0.5em}\begin{equation*}
\left\{
\begin{array}{cc}
    \underline{x}_{k+1,j}\leq \overline{x}_{k+1,j} \leq \underline{x}_{u,j} & \text{if } \delta^u_{1,j} = 1,\; \delta^u_{2,j} = 0\\
    \overline{x}_{k+1,j}\geq\underline{x}_{k+1,j} \geq \overline{x}_{u,j} & \text{if } \delta^u_{1,j} = 0,\; \delta^u_{2,j} = 1\\
    \underline{x}_{u,j}\leq\overline{x}_{k+1,j}\leq\overline{x},\; \underline{x}\leq\underline{x}_{k+1,j}\leq\overline{x}_{u,j} & \text{if } \delta^u_{1,j} = 0,\; \delta^u_{2,j} = 0
\end{array}
\right.,
\vspace{-0.5em}\end{equation*}
where $\overline{x}_{k+1,j},\underline{x}_{k+1,j},\overline{x}_{u,j},\underline{x}_{u,j}$ are the $j${th} element of $\overline{x}_{k+1},\underline{x}_{k+1},\overline{x}_{u},\underline{x}_{u}$, respectively. The integer constraints imply that there exists at least one dimension $j$ where $[\underline{x}_{k+1,j},\overline{x}_{k+1,j}]\cap[\underline{x}_{u,j},\overline{x}_{u,j}]=\varnothing$. Subsequently, the hypercube $[\underline{x}_{k+1},\overline{x}_{k+1}]$ has zero overlaps with the unsafe subset $[\underline{x}_{u},\overline{x}_{u}]$. Notably, in the case of a complex unsafe region $\mathcal{X}_u$, we can derive a union of hypercubes $\cup_i \mathcal{X}_u^{(i)}$ as in \eqref{eq:x_unsafe_set} that over-bounds $\mathcal{X}_u$. Thereafter, to ensure safety, we can formulate similar constraints \eqref{eq:mip_out_unsafe} with each individual hypercube $\mathcal{X}_u^{(i)}$ in the union. 
\if\arxiv1
The proof is available in Appendix~\ref{subsec:safety_proof_output}.
\fi

\subsection{Safe Tracking Control using MILP}\label{subsec:constraint_obj}
The optimization objective in \eqref{eq:OP_objective} is designed to minimize the maximum distance between the points in the reachable set and the reference state $x^r$. Focusing on the case when $p=1$, i.e., the cost is defined using $\ell_1$ norm, we can introduce a vector of slack variables $\lambda\in\R^{n_x}$, and reformulate \eqref{eq:OP_objective} as
\vspace{-0.5em}\begin{equation*}
\begin{array}{c}
    \argmin_{\tilde{u}_k} \sum_{q=1}^{n_x} \lambda_q,
\end{array}
\vspace{-0.5em}\end{equation*}
\begin{equation}\label{eq:mip_objective}
\text{subject to:\quad}
\begin{array}{c}
    \lambda\geq 0,\;
    -\lambda\leq \underline{x}_{k+1} - x^r \leq \lambda,\;
    -\lambda\leq \overline{x}_{k+1} - x^r \leq \lambda,
\end{array}
\end{equation}
where $\lambda_q$ designates the $q${th} element of vector $\lambda$. This objective function relies on the fact that the maximum distance, between a reference $x^r$ and points in the hypercube $[\underline{x}_{k+1}, \overline{x}_{k+1}]$, is attained at the points located at the boundary of the hypercube. Then, the optimization problem \eqref{eq:safeOptProblem} for tracking the reference state $x^r$ safely can be rewritten into a MILP according to

\problem{Robust Constrained Tracking Control Problem}
\vspace{-1em}\begin{equation}
\begin{array}{c}\label{eq:mip}
    \argmin\limits_{\substack{
    \tilde{u}_k,\; \delta^a,\; \delta^b,\; a_0,\; b_0,\; a_{k+1},\; b_{k+1},\; \underline{x}_{k+1},\; \overline{x}_{k+1},\; \delta^u_1,\; \delta^u_2,\; \lambda,\\
    a_i,\; b_i,\; \hat{a}_i,\; \hat{b}_i,\; \delta_i^{--},\; \delta_i^{-+},\; \delta_i^{++},\; i=1,\dots,\ell-1,
    }}\quad
    \sum\limits_{q=1}^{n_x} \lambda_q,\\
    \text{subject to: \eqref{eq:mip_input_x}, \eqref{eq:mip_input_u}, \eqref{eq:mip_struct_linear}, \eqref{eq:mip_struct_relu}, \eqref{eq:mip_struct_out}, \eqref{eq:mip_actual_reachable}, \eqref{eq:mip_out_unsafe}, \eqref{eq:mip_objective}.}
\end{array}    
\vspace{-0.5em}\end{equation}
We note that the number of decision variables in the MILP problem \eqref{eq:mip} scales linearly with the number of neurons in the NN. In the subsequent examples, we use the \textbf{\it YALMIP} toolbox (\cite{lofberg2004yalmip}) for MATLAB to solve the optimization \eqref{eq:mip}. The code is available at \url{https://github.com/XiaoLiSean/MILPSafetyGuard}. It also has the following property.  
\if\arxiv1
The proof is presented in Appendix~\ref{sec:safety_proof}.
\fi 

\begin{proposition}\label{prop:safety}
Consider a NN-learned dynamical system defined by \eqref{eq:nn_structure}, \eqref{eq:relu} that takes control $u_k$ and state $x_k$ as inputs and yields $\tilde{x}_{k+1}$ as a prediction of the next state $x_{k+1}$ and satisfies the bounded additive error assumption in \eqref{eq:x_bound}, \eqref{eq:x_bound_cube}. We assume that $x_k\in\mathcal{X}_s$ is unknown but belongs to a safe set $\mathcal{X}_s=\mathcal{X}\backslash\mathcal{X}_u$ that is the complement set of the unsafe region $\mathcal{X}_u$ given by \eqref{eq:x_unsafe_set} in the state space $\mathcal{X}$ defined by \eqref{eq:x_feasible_set}.  Also assume that a measurement $y_k$ of $x_k$ is given that satisfies the assumptions in \eqref{eq:y_bound}, \eqref{eq:y_bound_cube}. If there exists a solution of the Problem \eqref{eq:mip} such that the corresponding $\tilde{u}_k$ is in the admissible set $\mathcal{U}$ defined by \eqref{eq:u_feasible_set}, then for all actuator disturbances $w_k^u$ in set $\mathcal{W}_u$ defined by \eqref{eq:u_bound_cube}, the actual control $u_k$ subject to this additive disturbance $w_k^u$ according to \eqref{eq:u_bound} renders the actual next system state safe, i.e.,  $x_{k+1}\in\mathcal{X}_s$.
\end{proposition}

\section{Obstacle Avoidance and Reachability-Guided RRT}\label{sec:resultRobot}
\begin{figure}[ht]
\begin{center}
\includegraphics[width=\linewidth]{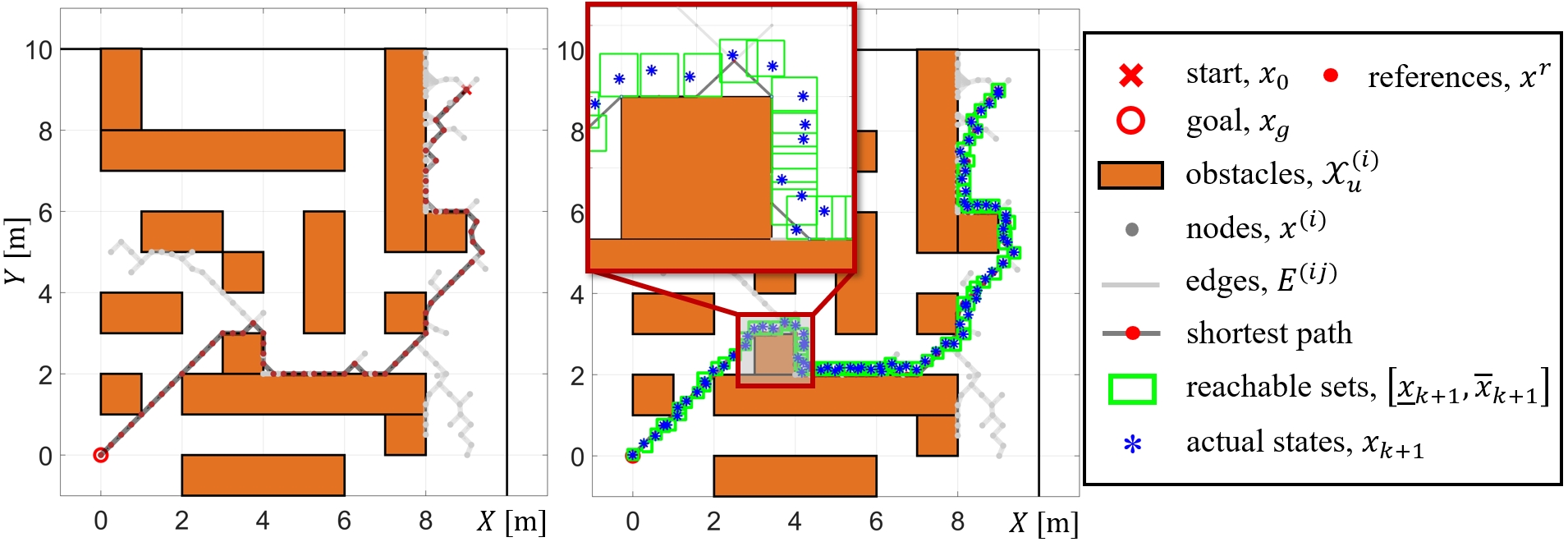}
\end{center}\vspace{-1.5em}
\caption{Schematic of obstacle avoidance using an omnidirectional robot: (Left) The reachability-guided RRT algorithm expands the tree from the start $x_0$ to the goal $x_g$ over the safe state space $\mathcal{X}_s$. Then, we use the Dijkstra planning algorithm to find a reference path (black lines with red dots) from $x_0$ to $x_g$ that has the shortest distance defined by $\ell_1$ norm. (Middle) We use the proposed method in \eqref{eq:mip} to track the reference states (red dots). Our method can guarantee that the robot motion is collision-free under uncertainties, i.e., the unknown actual states in blue asterisks are in the safe set $\mathcal{X}_s$, even with most of the reference points located near the obstacles.}
\label{fig:resultsRob}\vspace{-1em}
\end{figure}

We consider an omnidirectional robot as a point mass and use the following equation to represent its kinematics:  $x_{k+1}=f(x_k,u_k)=x_k+u_k + w_k$, where $x_k\in\mathbb{R}^2$ is the robot coordinates in the $X-Y$ plane, $u_k\in\mathbb{R}^2$ contains the displacements, and the values of the disturbance $w_k\sim U(-\epsilon^x, \epsilon^x)$ are sampled uniformly from the interval $[-\epsilon^x, \epsilon^x]$. We consider the sets $\mathcal{U} = \left\{[u_1\;u_2]^T\in\mathbb{R}^2: -0.25\leq u_1, u_2\leq -0.25\right\}$, $\mathcal{X} = \left\{[x_1\;x_2]^T\in\mathbb{R}^2: -1\leq x_1, x_2\leq 10\right\}$ together with obstacles $\mathcal{X}_u^{(i)}$ visualized as orange boxes in Figure~\ref{fig:resultsRob}. The NN is manually constructed and admits the following form
\vspace{-0.5em}\begin{equation*}
    \tilde{x}_{k+1} = 
    \left[
    \begin{array}{cc}
    -I_{2\times2} & -I_{2\times2}
    \end{array}
    \right]
    {\tt ReLU}
    \left(
    \left[
    \begin{array}{cc}
        I_{2\times2} & \zero_{2\times2}\\
        \zero_{2\times2} & I_{2\times2}
    \end{array}
    \right]
    \left[
    \begin{array}{c}
        x_k\\
        u_k
    \end{array}
    \right]
    +
    \left[
    \begin{array}{c}
        -50\cdot \one_{2\times 1}\\
        -50\cdot \one_{2\times 1}
    \end{array}
    \right]
    \right)
    +
    100\cdot \one_{2\times 1},
\vspace{-0.5em}\end{equation*}
which is equivalent to $\tilde{x}_{k+1} = \tilde{f}(x_k,u_k) = x_k + u_k$ given $x_k\in\mathcal{X}$ and  $u_k\in\mathcal{U}$. Finally, the uncertainty bounds are set to $\epsilon^x= \epsilon^u= \epsilon^y=[0.05,\;0.05]^T$. We develop a reachability-guided RRT planner similar to \cite{shkolnik2009reachability} using the CORA toolbox (\cite{CORA}), combined with the Dijkstra algorithm, to generate a path of reference states $x^r$. Different from the classic RRT, the reachability-guided RRT incorporates dynamics as a constraint to extend the edges of the tree. This tree $\mathcal{T}(\{x^{(i)}\},\{E^{(ij)}\})$ comprises nodes $x^{(i)}\in\mathcal{X}_s$ and directed edges $E^{(ij)}$. The edge $E^{(ij)}$ connects node $x^{(i)}$ to node $x^{(j)}$ and implies that there exists a control $u_k\in \mathcal{U}$ such that $x^{(j)} = \tilde{f}(x^{(i)}, u_k) \in \mathcal{X}_s$, i.e., $x^{(j)}\in \tilde{\mathcal{F}}(\{x^{(i)}\},\mathcal{U})\cap\mathcal{X}_s$. As shown in Figure~\ref{fig:resultsRob}, the algorithm is initialized with an initial node $x_0$ and is terminated if there exists a node $x^{(i)}\in \mathcal{T}$ such that $x_g\in \tilde{\mathcal{F}}(\{x^{(i)}\},\mathcal{U})$. At each time step, we take the first node $x^{(i)}$ in the shortest path as the reference state $x^r$ in \eqref{eq:mip}. We solve the optimization \eqref{eq:mip} and apply the resulting control $\tilde{u}_k$ to the actual dynamic model $f$. Then, we remove $x^{(i)}$ from the path if $x^{(i)}\in[\underline{x}_{k+1},\overline{x}_{k+1}]$ and the navigation terminates when $x_g\in[\underline{x}_{k+1},\overline{x}_{k+1}]$. The solution of \eqref{eq:mip} takes $0.16\pm 0.08$ sec on a laptop with an Intel i7-8550U CPU and 8 GB memory. As shown in Figure~\ref{fig:resultsRob}, after taking control $\tilde{u}_k$, the resulting next states $x_{k+1}$ in blue asterisks are within the reachable set $[\underline{x}_{k+1},\overline{x}_{k+1}]$ in green boxes and the boxes $[\underline{x}_{k+1},\overline{x}_{k+1}]$ are collision-free, which empirically validates Proposition~\ref{prop:safety}.
\section{Vehicle Navigation and Set-theoretical Localization}\label{sec:resultCar}
\begin{figure}[ht!]
\begin{center}
\includegraphics[width=\linewidth]{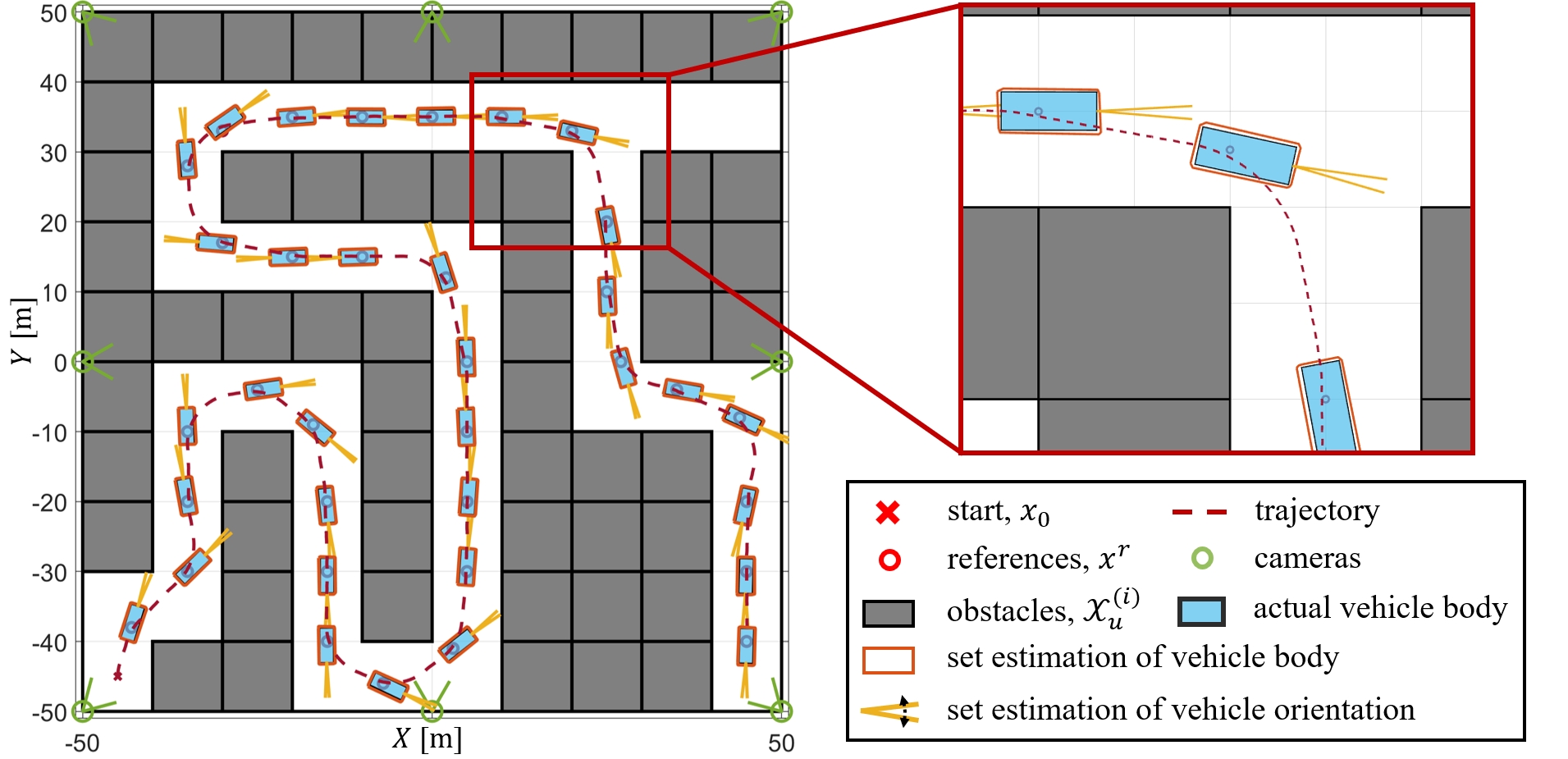}
\end{center}\vspace{-1.5em}
\caption{Schematic of navigating a vehicle through a maze with tracking controller \eqref{eq:mip} and with a set-theoretic localization algorithm. The zoom-in view, at the top-left, demonstrates that the set-theoretic localization algorithm provides estimates of the vehicle body and orientation that are guaranteed to contain the actual ones. The tracking controller \eqref{eq:mip} leverages this information, together with an NN-learned vehicle dynamics model, to avoid obstacles.}
\label{fig:resultsCar}\vspace{-1em}
\end{figure}

We next consider a front-wheel drive vehicle of $2\;\rm m$ width shown in Figure~\ref{fig:resultsCar}. The length of the vehicle wheelbase is $l=5\;\rm m$. We adopt the vehicle kinematics model from \cite{li2023set}, which admits the form,
\vspace{-0.5em}\begin{equation*}
x_{k+1} = f(x_k, u_k)
=
\left[
\begin{array}{c}
     p_{x, k} + v_k  dt \cos{\theta_k}
     \cos{\delta_k}  
     \\
      p_{x, k} + v_k  dt \sin{\theta_k} \cos{\delta_k}
     \\
     \theta_k + v_k dt /l \sin{\delta_k}
\end{array}
\right],
\vspace{-0.5em}\end{equation*}
where $x_k = [p_{x, k}\; p_{y, k}\; \theta_k]^T$ is the state vector, $(p_{x, k},\; p_{y, k})$ in meters are the coordinates of the center of the vehicle rear wheel axis, and $\theta_k\in [-\pi, \pi]$ is the vehicle orientation; $u_k = [v_k\;\delta_k]^T$ is the control vector, $v_k$ in $\rm m/s$ is the vehicle longitudinal speed, and $\delta_k$ in $\rm rad$ is the vehicle steering angle;  $dt=0.1\;\rm sec$ is the sampling period. We use the sets $\mathcal{U} = \left\{u_k: v_k\in[2, 5],\; \delta_k\in[-0.6, 0.6]\right\}$, $\mathcal{X} = \left\{x_k:  p_{x, k}, p_{y, k}\in[-50, 50],\; \theta_k\in [-\pi, \pi]\right\}$ together with obstacles $\mathcal{X}_u^{(i)}$ visualized in grey boxes in Figure~\ref{fig:resultsCar}. For the uncertainties, we assume the actuator disturbance $\epsilon^u = [0.01\;\rm m/s,\; 0.5 \;\rm deg]^T$. We densely sample a dataset $\mathcal{D}=\{(x^{(i)}_k, u^{(i)}_k, x^{(i)}_{k+1})\}_i$ from $\mathcal{X}\times \mathcal{U}$ for NN training and quantifying the NN prediction error. The NN has two hidden layers of 8 and 4 neurons, respectively. Based on Pytorch library (\cite{pytorch}), we train an NN $\tilde{f}$ using the Stochastic Gradient Descent algorithm and dataset $\mathcal{D}$ to minimize the mean-squared error $\norm{x^{(i)}_{k+1} - \tilde{f}(x^{(i)}_k, u^{(i)}_k)}^2_2$, and the prediction error is equal to $\epsilon^x = [0.02\;\rm m,\; 0.02\;\rm m,\; 1.5\;\rm deg]^T$. We quantify the prediction error as the maximum value of the empirical absolute error, i.e., 
\vspace{-0.5em}\begin{equation*}
    \epsilon^x = \max\left\{\epsilon\in\R^3: \epsilon = \abs{x^{(i)}_{k+1} - \tilde{f}(x^{(i)}_k, u^{(i)}_k)},\; (x^{(i)}_k, u^{(i)}_k, x^{(i)}_{k+1})\in\mathcal{D}\right\},
\vspace{-0.5em}\end{equation*}
and the function $\max$ is applied element-wise. The theoretical properties of uncertainty-bound quantification from samples are discussed in~\cite{dean2020robust} and are beyond the scope of this work. At each time step $k$, we apply the set-theoretic localization algorithm presented in \cite{li2023set} to generate an uncertainty polygon $P_{xy}$ that contains the actual vehicle position $[p_{x, k},\; p_{y, k}]^T$ and an uncertainty interval $P_{\theta}$ that contains the actual vehicle orientation $\theta_k$, i.e., $[p_{x, k},\; p_{y, k}]^T\in P_{xy}$ and $\theta_k\in P_{\theta}$. Subsequently, we can derive the smallest hypercube $P$ that over-bounds $P_{xy}\times P_{\theta}$, and the measurement error $\epsilon^y$ is quantified as half of the sizes of $P$. The algorithm can also produce a polytope estimation of the vehicle body as demonstrated in Figure~\ref{fig:resultsCar}. Then, akin to the process in Section~\ref{sec:resultRobot}, we solve the optimization problem \eqref{eq:mip} and use the resulting control to navigate the vehicle through the maze. The solution of \eqref{eq:mip} takes $0.14\pm0.06$ sec. As shown in Figure~\ref{fig:resultsCar}, we can also observe that the vehicle motion is collision-free, which is consistent with Proposition~\ref{prop:safety}. 
\section{Conclusion and Future Work}\label{sec:conclusion}
In this paper, we developed an approach for robust reference tracking that leveraged a learned NN model to control the actual dynamics. We considered both bounded intrinsic and extrinsic uncertainties from the controller and other system modules, respectively. We transcripted the resulting decision-variable-dependent uncertainty set propagation through NN using a MILP. We provided results which ensure that the proposed MILP can render the overall system safe considering all possible actuator disturbances, measurement noise, and prediction errors within their corresponding bounded sets. We tested the proposed method in navigation and obstacle avoidance scenarios for an omnidirectional robot and a vehicle in simulations. We also note that the recursive feasibility is not guaranteed under the current horizon-one MPC setting; this will be addressed in future work.


\if\arxiv1
\bibliography{ref}
\else
\newpage
\bibliography{ref}
\fi 
\if\arxiv1
\newpage
\appendix
\section{Formal Safety Guarantee}\label{sec:safety_proof}
The proofs of Proposition~\ref{prop:constraint_input},~\ref{prop:constraint_nn},~\ref{prop:constraint_output} are presented in~\ref{subsec:safety_proof_input},~\ref{subsec:safety_proof_nn},~\ref{subsec:safety_proof_output}, respectively. Subsequently, the hypercube defined by the lower and upper bounds $\underline{x}_{k+1},\overline{x}_{k+1}$, as a result of the MILP problem \eqref{eq:mip}, contains the actual state $x_{k+1}$, and the hypercube is within the safe set $\mathcal{X}_s$, i.e., $x_{k+1}\in[\underline{x}_{k+1},\overline{x}_{k+1}]\subset\mathcal{X}_s$. This proves Proposition~\ref{prop:safety}.

\subsection{Input Constraints}\label{subsec:safety_proof_input}
\begin{proof}
Based on assumption in \eqref{eq:y_bound} and $x_k\in \mathcal{X}_s\subset \mathcal{X}$, it's obvious that $x_k\in \mathcal{X}_k=(y_k\oplus-\mathcal{W}_y)\cap\mathcal{X}_s$ by the definition of Minkowski Sum, thereby we have $x_k\in \mathcal{X}_k\subset (y_k\oplus-\mathcal{W}_y)\cap\mathcal{X}$. As defined in \eqref{eq:mip_input_x} together with the hypercube assumptions in \eqref{eq:y_bound_cube}, \eqref{eq:x_feasible_set}, the following statement can be proved 
\begin{equation*}
    \forall x\in (y_k\oplus-\mathcal{W}_y)\cap\mathcal{X},\;a_{0,1: n_x} = \max\{\underline{x}, y_k-\epsilon^y\} \leq x \leq \min\{\overline{x}, y_k+\epsilon^y\} = b_{0,1: n_x}.
\end{equation*}
Therefore, we can conclude that 
\begin{equation}\label{eq:proof_input_x}
    x_k\in\left[a_{0,1: n_x},\; b_{0,1: n_x}\right].
\end{equation}
Similarly, based on \eqref{eq:u_bound} and $u_k\in\mathcal{U}$, we can show that $u_k\in\mathcal{U}_k$ and $\max\{\underline{u}, \tilde{u}_k-\epsilon^u\} \leq u \leq \min\{\overline{u}, \tilde{u}_k+\epsilon^u\}$ for all $u\in\mathcal{U}_k$. Next, we need to show that the constraints \eqref{eq:mip_input_u} imply $a_{0,(n_x+1): (n_x+n_u)} = \max\{\underline{u}, \tilde{u}_k-\epsilon^u\}$ and $b_{0,(n_x+1): (n_x+n_u)}= \min\{\overline{u}, \tilde{u}_k+\epsilon^u\}$ from which we can conclude 
\begin{equation}\label{eq:proof_input_u}
    u_k\in\left[a_{0,(n_x+1): (n_x+n_u)},\; b_{0,(n_x+1): (n_x+n_u)}\right].
\end{equation}

If $\delta^a_j=1$, the constraints \eqref{eq:mip_input_u} yield the following inequalities
\begin{equation*}
\left\{\begin{array}{l}
    a_{0,j} \geq \underline{u}_j\\
    a_{0,j} \geq \tilde{u}_{k,j}-\epsilon^u_j\\
    a_{0,j} \leq \underline{u}_j\\
    a_{0,j} \leq \tilde{u}_{k,j}-\epsilon^u_j + \max\{\epsilon^u_j, \overline{u}_j-\underline{u}_j-\epsilon^u_j\}
\end{array}\right.
\end{equation*}
where the first three inequalities are equivalent to $a_{0,j}=\underline{u}_j,\; \underline{u}_j \geq \tilde{u}_{k,j}-\epsilon^u_j$. In this case, we have $a_{0,j}= \max\{\underline{u}_j, \tilde{u}_{k,j}-\epsilon^u_j\}$ and the fourth inequality is valid since $\underline{u}_j\leq \tilde{u}_{k,j}\leq\overline{u}_j$ and $a_{0,j}-\tilde{u}_{k,j}+\epsilon^u_j\leq\underline{u}_j-\underline{u}_j+\epsilon^u_j \leq  \epsilon^u_j \leq \max\{\epsilon^u_j, \overline{u}_j-\underline{u}_j-\epsilon^u_j\}$.

If $\delta^a_j=0$, the constraints \eqref{eq:mip_input_u} produce the following inequalities
\begin{equation*}
\left\{\begin{array}{l}
    a_{0,j} \geq \underline{u}_j\\
    a_{0,j} \geq \tilde{u}_{k,j}-\epsilon^u_j\\
    a_{0,j} \leq \tilde{u}_{k,j}-\epsilon^u_j\\
    a_{0,j} \leq \underline{u}_j + \max\{\epsilon^u_j, \overline{u}_j-\underline{u}_j-\epsilon^u_j\}
\end{array}\right.
\end{equation*}
where the first three inequalities are equivalent to $a_{0,j}=\tilde{u}_{k,j}-\epsilon^u_j,\; \tilde{u}_{k,j}-\epsilon^u_j \geq \underline{u}_j$. Again, we can show $a_{0,j}= \max\{\underline{u}_j, \tilde{u}_{k,j}-\epsilon^u_j\}$ and the fourth inequality is also feasible since $a_{0,j}-\underline{u}_j=\tilde{u}_{k,j}-\epsilon^u_j-\underline{u}_j \leq  \overline{u}_j-\epsilon^u_j-\underline{u}_j \leq \max\{\epsilon^u_j, \overline{u}_j-\underline{u}_j-\epsilon^u_j\}$. 

Hence, we can conclude that $a_{0,j}= \max\{\underline{u}_j, \tilde{u}_{k,j}-\epsilon^u_j\}$ for $j=(n_x+1),\dots,(n_x+n_u)$ that is equivalent to $a_{0,(n_x+1): (n_x+n_u)} = \max\{\underline{u}, \tilde{u}_k-\epsilon^u\}$ and the proof of $b_{0,(n_x+1): (n_x+n_u)}= \min\{\overline{u}, \tilde{u}_k+\epsilon^u\}$ resembles the discussion above. Eventually, based on \eqref{eq:proof_input_x} and \eqref{eq:proof_input_u}, the input $z_0$ to the NN satisfies
\begin{equation}\label{eq:proof_input}
    z_0=[x_k^T\; u_k^T]^T\in\left[a_0,\;b_0\right]. 
\end{equation}
\end{proof}

\subsection{NN Structural Constraints}\label{subsec:safety_proof_nn}
\begin{proof}
We first provide proof of the following statement
\begin{equation*}
    \hat{z}_i = W^{(i)}z_{i-1} +b^{(i)}\in \left[\hat{a}_i,\;\hat{b}_i\right],\;
    z_i = \max\{0,\;\hat{z}_i\}\in \left[a_i,\;b_i\right],\; 
    \text{given }
    z_{i-1}\in \left[a_{i-1},\;b_{i-1}\right].
\end{equation*}
Then, from the results $z_0\in\left[a_0,\;b_0\right]$ in \eqref{eq:proof_input}, we can inductively prove the same argument for $i=1,\dots,\ell-1$. 

Given $z_{i-1}\in \left[a_{i-1},\;b_{i-1}\right]$, the $j${th} element in $\hat{z}_i$ is $\hat{z}_{i,j}=w^{(i)}_jz_{i-1}+b^{(i)}_j$ where $w^{(i)}_j=[w_1\dots w_q\dots w_{n_{i-1}}]$ and satisfies the following inequalities
\begin{equation*}
\begin{array}{c}
    \sum\limits_{q=1}^{n_{i-1}} \left( \IversonBracket{w_q\geq0}\cdot\left(w_q a_{i-1,q}\right) + \IversonBracket{w_q<0}\cdot\left(w_q b_{i-1,q}\right)\right)
    \\
    \leq \hat{z}_{i,j} - b^{(i)}_j \leq 
    \\
    \sum\limits_{q=1}^{n_{i-1}} \left( \IversonBracket{w_q\geq0}\cdot\left(w_q b_{i-1,q}\right) + \IversonBracket{w_q<0}\cdot\left(w_q a_{i-1,q}\right)\right),
\end{array}
\end{equation*}
where the Iverson bracket $\IversonBracket{\cdot}$ takes the value of 1 if the statement inside is true and 0 otherwise. We can derive the following equalities from the constraints in \eqref{eq:mip_struct_linear}
\begin{equation*}
\begin{aligned}
    \hat{a}_{i,j} - b^{(i)}_j &= \sum\limits_{q=1}^{n_{i-1}} \left( \IversonBracket{w_q\geq0}\cdot\left(w_q a_{i-1,q}\right) + \IversonBracket{w_q<0}\cdot\left(w_q b_{i-1,q}\right)\right),\\
    \hat{b}_{i,j} - b^{(i)}_j &= \sum\limits_{q=1}^{n_{i-1}} \left( \IversonBracket{w_q\geq0}\cdot\left(w_q b_{i-1,q}\right) + \IversonBracket{w_q<0}\cdot\left(w_q a_{i-1,q}\right)\right),
\end{aligned}
\end{equation*}
which implies
\begin{equation}\label{eq:proof_struct_linear}
    \hat{z}_{i,j}\in\left[\hat{a}_{i,j},\;\hat{b}_{i,j}\right],\; j=1,\dots,n_i.
\end{equation}

Afterward, given $\underline{\hat{z}}_{i,j}\leq\hat{a}_{i,j}\leq\hat{z}_{i,j}\leq \hat{b}_{i,j}\leq \overline{\hat{z}}_{i,j}$ in \eqref{eq:mip_struct_linear} and constraints in \eqref{eq:mip_struct_relu}, we need to prove $a_{i,j}\leq z_{i,j}\leq b_{i,j}$. If $\delta^{--}_{i,j}=1$, the constraints in \eqref{eq:mip_struct_relu} are reduced to 
\begin{equation*}
\left\{\begin{array}{l}
    b_{i,j}\geq a_{i,j} \geq 0\\
    \hat{a}_{i,j} \leq a_{i,j} \leq 0\\
    \hat{b}_{i,j} \leq b_{i,j} \leq 0\\
    a_{i,j} \leq \hat{a}_{i,j} -\underline{\hat{z}}_{i,j}\\
    b_{i,j} \leq \hat{b}_{i,j} - \underline{\hat{z}}_{i,j}
\end{array}\right..
\end{equation*}
The first three inequalities are equivalent to $a_{i,j}=b_{i,j}=0$, $ \hat{a}_{i,j}\leq 0$, $\hat{b}_{i,j}\leq 0$ which induce $\hat{z}_{i,j}\leq \hat{b}_{i,j}\leq 0$ and $z_{i,j} = \max\{0,\;\hat{z}_{i,j}\}=0$, thereby we have $0=a_{i,j}\leq z_{i,j}\leq b_{i,j}=0$. Meanwhile, the last two inequalities hold since $\underline{\hat{z}}_{i,j}\leq\hat{a}_{i,j}\leq\hat{z}_{i,j}\leq \hat{b}_{i,j}\leq \overline{\hat{z}}_{i,j}$. 

If $\delta^{-+}_{i,j}=1$, the constraints in \eqref{eq:mip_struct_relu} are reduced to 
\begin{equation*}
\left\{\begin{array}{l}
    b_{i,j}\geq a_{i,j} \geq 0\\
    \hat{a}_{i,j} \leq a_{i,j} \leq 0\\
    \hat{b}_{i,j} \leq b_{i,j} \leq \hat{b}_{i,j}\\
    a_{i,j} \leq \hat{a}_{i,j} -\underline{\hat{z}}_{i,j}\\
    b_{i,j} \leq \overline{\hat{z}}_{i,j}
\end{array}\right..
\end{equation*}
The first three inequalities imply that $a_{i,j}=0$, $\hat{a}_{i,j}\leq 0$ and $b_{i,j}=\hat{b}_{i,j}$, $\hat{b}_{i,j}\geq 0$ from which we can show that $0\leq z_{i,j}\leq \hat{b}_{i,j}$. Thus, we have $a_{i,j}=0\leq z_{i,j}\leq \hat{b}_{i,j} = b_{i,j}$ and the last two inequalities are also feasible.

If $\delta^{++}_{i,j}=1$, the constraints in \eqref{eq:mip_struct_relu} are reduced to 
\begin{equation*}
\left\{\begin{array}{l}
    b_{i,j}\geq a_{i,j} \geq 0\\
    \hat{a}_{i,j} \leq a_{i,j} \leq \hat{a}_{i,j}\\
    \hat{b}_{i,j} \leq b_{i,j} \leq \hat{b}_{i,j}\\
    a_{i,j} \leq \overline{\hat{z}}_{i,j}\\
    b_{i,j} \leq \overline{\hat{z}}_{i,j}
\end{array}\right..
\end{equation*}
The first three inequalities imply that $a_{i,j}=\hat{a}_{i,j}$, $\hat{a}_{i,j}\geq 0$ and $b_{i,j}=\hat{b}_{i,j}$, $\hat{b}_{i,j}\geq 0$ from which we can show $\hat{a}_{i,j}\leq z_{i,j}\leq \hat{b}_{i,j}$. Thus, we have $a_{i,j}=\hat{a}_{i,j}\leq z_{i,j}\leq \hat{b}_{i,j} = b_{i,j}$ and the last two inequalities are valid. To this end, we have proven that
\begin{equation}\label{eq:proof_struct_relu}
    z_{i,j}\in\left[a_{i,j},\;b_{i,j}\right],\; j=1,\dots,n_i.
\end{equation}
Indeed, as discussed at the beginning of this section, we can show that
\begin{equation}\label{eq:proof_struct}
    \hat{z}_i \in \left[\hat{a}_i,\;\hat{b}_i\right],\;
    z_i \in \left[a_i,\;b_i\right],\; i=1,\dots,\ell-1
\end{equation}
with \eqref{eq:proof_input}. Identical to the proof of \eqref{eq:proof_struct_linear}, given $z_{\ell-1} \in \left[a_{\ell-1},\;b_{\ell-1}\right]$ in \eqref{eq:proof_struct}, constraints in \eqref{eq:mip_struct_out} and $\tilde{x}_{k+1} = W^{(\ell)}z_{\ell-1} +b^{(\ell)}$ in \eqref{eq:nn_structure}, we can demonstrate that
\begin{equation}\label{eq:proof_output_tilde}
    \tilde{x}_{k+1}\in \tilde{\mathcal{F}}(\mathcal{X}_k,\mathcal{U}_k)\subseteq \left[a_{k+1},\;b_{k+1}\right].
\end{equation}
\end{proof}

\subsection{Output Constraints}\label{subsec:safety_proof_output}
\begin{proof}
With the assumptions in \eqref{eq:x_bound} and \eqref{eq:x_bound_cube}, we can show $x_{k+1}\in \left[a_{k+1},\;b_{k+1}\right] \oplus \mathcal{W}_x$ which can be further derived to
\begin{equation}\label{eq:proof_output_xkp1}
    x_{k+1}\in \left[\underline{x}_{k+1},\; \overline{x}_{k+1}\right]\subset \mathcal{X},
\end{equation}
based on the constraints in \eqref{eq:mip_actual_reachable}. Meanwhile, the constraints in \eqref{eq:mip_out_unsafe} are equivalent to the following statement
\begin{equation*}
    \exists j\in \{1,\dots, n_x\},\; s.t.\; \delta^u_{1,j}=1 \text{ or } \delta^u_{2,j}=1.
\end{equation*}

In the sequel, we show that $\left[\underline{x}_{k+1},\; \overline{x}_{k+1}\right]\cap \mathcal{X}_u=\varnothing$ with constraints in \eqref{eq:mip_out_unsafe} if $\delta^u_{1,j}=1$ which the proof resembles in case of $\delta^u_{2,j}=1$. If $\delta^u_{1,j}=1$, the constraints in \eqref{eq:mip_out_unsafe} can be reformulated into
\begin{equation*}
    \underline{x}_j\leq \underline{x}_{k+1,j}\leq \overline{x}_{k+1,j} \leq \underline{x}_{u,j}.
\end{equation*}
Suppose $\left[\underline{x}_{k+1},\; \overline{x}_{k+1}\right]\cap \mathcal{X}_u\neq\varnothing$, therefore there exists $x^*=[x^*_1\cdots x^*_j\cdots x^*_{n_x}]^T\in\mathbb{R}^{n_x}$ such that $x^*\in\left[\underline{x}_{k+1},\; \overline{x}_{k+1}\right]\cap \mathcal{X}_u$. Then, for some index $j$ and $1\leq j\leq n_x$, we have $\underline{x}_{u,j}\leq x^*_j\leq \overline{x}_{u,j}$ and $\underline{x}_{k+1,j}\leq x^*_j\leq \overline{x}_{k+1,j}$ which violates $\overline{x}_{k+1,j} \leq \underline{x}_{u,j}$. By contradiction, the following is true
\begin{equation}\label{eq:proof_output_unsafe}
    \left[\underline{x}_{k+1},\; \overline{x}_{k+1}\right]\cap \mathcal{X}_u=\varnothing.
\end{equation}
With equations \eqref{eq:proof_output_xkp1} and \eqref{eq:proof_output_unsafe}, we demonstrate that $x_{k+1}\in\mathcal{F}(\mathcal{X}_k,\mathcal{U}_k) \subseteq [\underline{x}_{k+1}, \overline{x}_{k+1}] \subset\mathcal{X}_s$ which complete the proof for Proposition~\ref{prop:constraint_output}.
\end{proof}
\fi 
\end{document}